# PSO-based Fuzzy Markup Language for Student Learning Performance Evaluation and Educational Application

Chang-Shing Lee, *Senior Member, IEEE*, Mei-Hui Wang, Chi-Shiang Wang
Olivier Teytaud, Jialin Liu, Su-Wei Lin, and Pi-Hsia Hung

*Abstract*—**Fuzzy relationships exist between students' learning performance with various abilities and a test item. However, the challenges in implementing adaptive assessment agents are obtaining sufficient items, efficient and accurate computerized estimation, and a substantial feedback agent. Additionally, the agent must immediately estimate students' ability item by item, which places a considerable burden on the server, especially for a group test. Hence, the implementation of adaptive assessment agent is more difficult in practice. This paper proposes an agent with particle swarm optimization (PSO) based on a Fuzzy Markup Language (FML) for students' learning performance evaluation and educational applications, and the proposed agent is according to the response data from a conventional test and an item response theory (IRT)-based three-parameter logistic (3PL) model. First, we apply a Gauss–Seidel (GS)-based parameter estimation mechanism to estimate the items' parameters according to the response data, and then to compare its results with those of an IRT-based Bayesian parameter estimation mechanism. In addition, we propose a static-IRT test assembly mechanism to assemble a form for the conventional test. The presented FML-based dynamic assessment mechanism infers the probability of making a correct response to the item for a student with various abilities. Moreover, this paper also proposes a novel PSO-based FML (PFML) learning mechanism for optimizing the parameters between items and students. Finally, we adopt a K-fold cross validation mechanism to evaluate the performance of the proposed agent. Experimental results show that the novel PFML learning mechanism for the parameter estimation and learning optimization performs favorably. We believe the proposed PFML will be a reference for education research and pedagogy and an important co-learning mechanism for future human–machine educational applications.**

## I. INTRODUCTION

Fuzzy Markup Language (FML) has been an IEEE 1855-2016 standard since May 2016. Based on XML, this language describes the knowledge base and rule base of a fuzzy logic system [1], [2], and features understandability, extendibility, and compatibility of implemented programs as well as programming efficiency [3]–[5]. FML facilitates the modelling of a fuzzy controller in a human-readable and hardware-independent manner [4]. Considerable research has focused on FML applications, including ambient intelligence frameworks [6], healthcare [7], computer games [8], and diet [9]. In addition, Lee *et al.* [10]–[12] used FML to represent the semantic structure of experts' knowledge about fuzzy ontology-based systems. Acampora *et al.* [13] proposed an FML script featuring evolving capabilities through a scripting language approach.

Inspired by the collective behavior of social animals, swarm-based algorithms have emerged as a powerful family of optimization techniques [14]. These algorithms have recently emerged as a family of nature-inspired, population-based algorithms capable of producing low-cost, fast, and robust solutions to various complex problems [14]. Particle swarm optimization (PSO) was successfully applied to neuro-fuzzy or neural network training [15]. Zhao *et al.* [16] proposed a self-adaptive harmony PSO search algorithm to solve global continuous optimization problems. Martinez-Soto *et al.* [17] proposed a hybrid PSO-GA optimization method for the automatic design of fuzzy logic controllers to minimize the steady state error of a plant's response. Songmuang and Ueno [18] proposed the bees algorithm for construction of multiple test forms in e-testing.

The current goals of education are not only to provide students with a complete education and a learning environment that enhances students' overall competitiveness but also to understand their learning performance and the conditions required to provide them with the appropriate guidance and pedagogy [19]. In item response theory (IRT)-based models [20], estimating the parameters of the items for a specific student group is a crucial task. The "*item*" of IRT denotes the "*question*" of the test paper. In a case where the items'

The authors would like to thank the financial support sponsored by 1) Ministry of Science and Technology (MOST) of Taiwan under the grant "INRIA-MOST Associate Team Program, 106-3114-E-024-001, 104-2221-E-024-015, and 105-2221-E-024-017" as well as 2) Kaohsiung City Government of Taiwan under the projects "Computerized Adaptive Assessment System Construction" and "Program of Learning Diagnosis and Progress Assessment for Primary and Secondary Students of Kaohsiung technology-based testing."

Chang-Shing Lee, Mei-Hui Wang, and Chi-Shiang Wang are with the Department of Computer Science and Information Engineering, National University of Tainan, Taiwan (e-mail: leecs@mail.nutn.edu.tw).

Olivier Teytaud and Jianlin Liu are with TAO, INRIA, University Paris-Sud 11, France.

Su-Wei Lin and Pi-Hsia Hung are with the Department of Education, National University of Tainan, Taiwan.

parameters are known, the maximum likelihood estimation method is a standard approach to estimating the students' abilities. The Bayesian estimation method in two-parameter logistic and three-parameter logistic (3PL) models estimates item parameters and students' abilities simultaneously [22]. An efficient assessment can identify a learner's strengths and weaknesses as well as what was learned and not learned [23]; therefore, creating an assembled test appropriate for the given examinees is crucial. Owing to the popularization of information technology and the Internet, computerized adaptive testing has become increasingly prevalent worldwide [24]. Accordingly, the aim of this paper is to propose a novel PSO-based FML (PFML) learning mechanism by integrating FML, PSO, and IRT for the dynamic assessment of student learning performance evaluation and educational applications. Studies [23]–[28] have discussed student learning performance evaluation and educational applications on the basis of IRT. However, no studies have combined FML with a PSO and IRT model or applied the new model to student learning performance evaluation.

Implementing a computerized adaptive assessment agent is challenging. First, a reliable, valid assessment must have sufficient items, which requires a considerable time and effort from domain experts. Second, the computational power of the server should be sufficiently efficient and accurate to handle the heavy burden of a mass group test, which means setting up a high-quality Internet environment that includes costly machines. Third, a feedback agent should be available for the students and teachers. To alleviate these challenges, this paper proposes a PSO system based on an FML for students' learning performance evaluation and educational application. The novelty and contribution of this paper are as follows: 1) This study is the first to apply IEEE 1855-2016 Standard FML [1], [2] and a PSO machine learning mechanism to an IRT model for student learning performance evaluation and educational applications. 2) This study combines IRT with an evolutionary strategy (ES) to assemble test papers for students with a specific range of abilities. Additionally, we combine IRT with the Gauss–Seidel (GS) method to estimate the parameters of the items. 3) We propose a novel human–machine co-learning model based on IRT and a human fuzzy linguistic knowledge cognition model for future educational applications. 4) Fuzzy logic and fuzzy sets with machine learning mechanisms are suitable for the evaluation of students' learning performance and educational applications. 5) The proposed agent is feasible for implementation in a large, computerized adaptive assessment into practice and as a reference for education research and practice.

The remainder of this paper is organized as follows: Section II briefly introduces the basic concepts of IRT, such as the item characteristic curve and test information. In addition, a novel static-IRT test assembly mechanism and a GS-based parameter estimation mechanism for educational dynamic assessment are presented. Section III describes the PFML optimization mechanism, including an FML-based dynamic assessment mechanism, a PFML learning mechanism, and a dynamic restriction mechanism for tuning the knowledge base of FML.

Section IV describes various experiments to evaluate validity of the proposed agent. Finally, Section V presents some conclusions and future educational applications by combining the proposed agent with artificial intelligence.

## II. STATIC-IRT TEST ASSEMBLY AND GAUSS-SEIDEL-BASED PARAMETER ESTIMATION FOR FML

In this section, we briefly introduce IRT, including the item characteristic curve (ICC), test characteristic curve, item information, test information, and test standard error (TSE). Next, we present a novel static-IRT test assembly mechanism and a novel GS-based parameter estimation mechanism for the PSO-based FML optimization of dynamic assessment.

### A. Item Characteristic Curve, Test Characteristic Curve, and Item Information

IRT features parameter invariance and an information function [20]. When the adopted IRT model perfectly fits the analyzed data, the distribution of students' abilities will not affect the results of the item parameter estimation and vice versa [20]. The four levels for student's performance defined in the IRT are below basic, basic, proficient, and advanced [19]. According to the 3PL model of dichotomous scoring, one item has three parameters, namely, $a$, $b$, and $c$, to represent this item's discrimination, difficulty, and guessing, respectively [20]. The ranges of $a$, $b$, and $c$ are defined as $[0, 2]$, $[-4, +4]$, and $[0, 1]$, respectively. The ICC describes the relationship between the ability of individuals and their probability of answering a test question correctly. The ICC for each item is calculated by (1). The $x$ and $y$ axes of the ICC denote the student's ability $\theta$ and probability $P(\theta)$ of providing the correct response to this item, respectively.

TABLE I
AN EXAMPLE OF PARAMETERS $A$, $B$, AND $C$ FOR 20 ITEMS

| PRM No. | $a$ | $b$ | $c$ | PRM No. | $a$ | $b$ | $c$ |
|---|---|---|---|---|---|---|---|
| 1 | 1.1 | 0 | 0.1 | 11 | 1 | -2 | 0 |
| 2 | 0.77 | 0.75 | 0.23 | 12 | 1 | 0 | 0.5 |
| 3 | 0.7 | -0.06 | 0.14 | 13 | 1 | 0 | 0.25 |
| 4 | 1.6 | 0 | 0.11 | 14 | 1 | 0 | 0 |
| 5 | 2 | 1.7 | 0.03 | 15 | 0.5 | -0.5 | 0.5 |
| 6 | 1.5 | 0 | 0 | 16 | 1 | 0 | 0.25 |
| 7 | 1 | 0 | 0 | 17 | 1.5 | 0.5 | 0 |
| 8 | 0.5 | 0 | 0 | 18 | 0.5 | 1 | 0 |
| 9 | 1 | 2 | 0 | 19 | 1 | 1.5 | 0.5 |
| 10 | 1 | 0 | 0 | 20 | 1.5 | -1 | 0.25 |

$$P(u_i = 1|\theta) = c_i + (1 - c_i) \times \frac{1}{1 + e^{-1.7 \times a_i(\theta - b_i)}} \quad (1)$$

where $u_i$ denotes the response data for the $i$th item; $\theta$ is the student's ability; and $a_i$, $b_i$, and $c_i$ are discrimination, difficulty, and guessing of the $i$th item, respectively.

Table I shows an example of $a$, $b$, and $c$ for 20 items. Figs. 1(a)–(c) show the effect on ICC for parameters $a$, $b$, and $c$, respectively. Fig. 1(a) reveals that the higher $a$ is, the steeper the ICC is. Fig. 1(b) indicates that $P(\theta) = 0.5$ when $\theta = b$. Fig. 1(c) shows that $P(\theta) = 0.5$ when $\theta = b$ and $c = 0$. Fig. 1(c) also demonstrates that $P(\theta) > 0.5$ when $\theta = b$ and $c > 0$. Fig. 2 presents the item information of items 15 to 20, with their parameters listed in Table I.

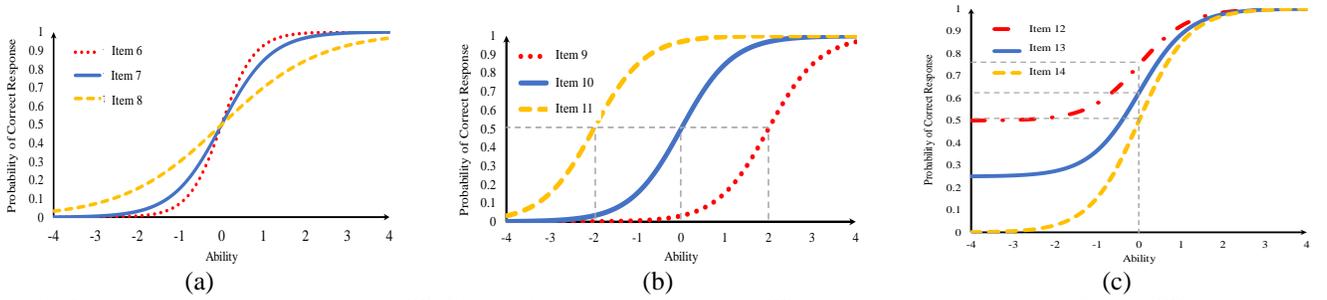

Fig. 1. (a) Discrimination (parameter $a$) effect on ICC, (b) difficulty (parameter $b$) effect on ICC, and (c) guessing (parameter $c$) effect on ICC.

The item information $I_i(\theta)$ of the $i^{th}$ item is calculated using (2). Fig. 2 shows that $I_i(\theta)$ decreases as the difference between the ability level and parameter $b$ increases, $I_i(\theta)$ approaches zero at the extremes of the ability scale, and the closer $b$ is to $\theta$, the higher $I_i(\theta)$ becomes. The higher the value of $I_i(\theta)$ is, the higher the value of the $i^{th}$ item becomes [20]. These results indicate that the $i^{th}$ item can provide useful information for the domain experts.

$$I_i(\theta) = 2.89 a_i^2 \left(\frac{Q_i(\theta)}{P_i(\theta)}\right)\left(\frac{P_i(\theta) - c_i}{1 - c_i}\right)^2, \text{ where } Q_i(\theta) = 1 - P_i(\theta) \quad (2)$$

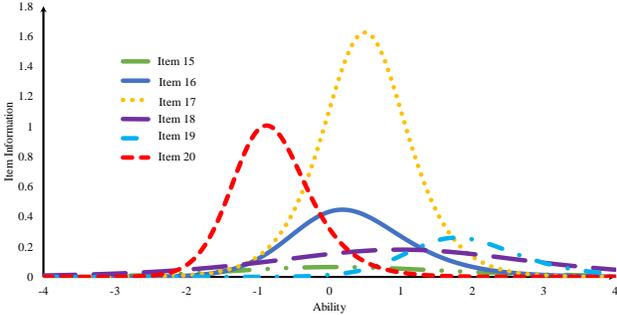

Fig. 2. Item information of Items 15 to 20.

### B. Test Information, Test Standard Error and IRT-based Bayesian Estimation Mechanism

The item information function and test information function (TIF) could be the basis for assembling the test paper during the test. The test information for a given ability level is the sum of the item information at that level, showing how well the test can estimate ability over the entire range of ability scores. When the TIF reaches its maximum, the test standard error (TSE) will reach its minimum value. The TIF and TSE are calculated by (3) and (4), respectively.

$$TIF(\theta) = \sum_{i=1}^{M} I_i(\theta) \quad (3)$$

where $M$ denotes the selected-item number and $\theta$ denotes the student's ability.

$$TSE(\theta) = \frac{1}{\sqrt{TIF(\theta)}} \quad (4)$$

When $TSE(\theta)$ is less than 0.3 the measurement precision for all of the students can reach the 0.9 reliability estimated by classical test theory [20]. Next, we briefly introduce the IRT-based Bayesian estimation (IRTBE) mechanism. After defining an items' parameters, we adopt the maximum *a posteriori* estimation approach to estimate a student's ability $\hat{\theta}$ based on their response pattern $\mathbf{u}$ to items 1, 2, 3, …, and $|I|$ [20]. By contrast, when the items parameters are unknown, the IRTBE mechanism can estimate them according to the examinees' abilities [29]. The inputs of IRTBE are items 1, 2, 3, ..., and $|I|$ and the response pattern $\mathbf{u} = (u_1, u_2, …, u_{|I|})$ for a student. The output is the estimated ability $\hat{\theta}$ for the student. We define the stages of IRTBE as follows:

1) **Stage 1**: Calculate the student's joint probability for their response to $I$ items.

$$P(u_1, u_2, u_3, …, u_{|I|}|\theta) = P(u_1|\theta) \times P(u_2|\theta) \times P(u_3|\theta)… \times P(u_{|I|}|\theta)$$

where

$$u_i = \begin{cases} 1, \text{ make a right response to the } i^{th} \text{ item} \\ 0, \text{ make a wrong response to the } i^{th} \text{ item} \end{cases}, \text{ and } 1 \le i \le |I|$$

2) **Stage 2**: Calculate the likelihood function $L(\mathbf{u}|\theta)$ according to response pattern $\mathbf{u}$, the probability of the correct response to the item $P$, the probability of an incorrect response to the item $Q$, and the student's ability $\theta$.

$$L(\mathbf{u}|\theta) = L(u_1, u_2, …, u_{|I|}|\theta) = \prod_{i=1}^{|I|} P(u_i|\theta) = \prod_{i=1}^{|I|} P_i^{u_i} Q_i^{1-u_i}$$

3) **Stage 3**: Calculate the posterior density function $f(\theta|\mathbf{u})$ based on prior density functions $f(\theta)$ and $L(\mathbf{u}|\theta)$.

$$f(\theta|\mathbf{u}) \propto L(\mathbf{u}|\theta) f(\theta)$$

4) **Stage 4**: Execute the search algorithm to find the $\theta$ that maximizes the posterior density function $(\hat{\theta})$.

$$\hat{\theta} = argmax \, f(\theta|\mathbf{u})$$

### C. Static-IRT Test Assembly Mechanism for Fuzzy Markup Language

To assemble the most suitable test paper for the specific students from a test item set, we combine IRT with ES [30, 31] to optimize the parameters of the objective function for the static-IRT test assembly mechanism. Table II shows the algorithm of the static-IRT test assembly mechanism for FML, and Fig. 3 shows its flowchart for optimizing the parameters of the objective function based (1+ 1)-ES. The vector $U$ shows the importance for all the items in set $I$ and σ is the global step size for the rate of successful mutations if $g(U') < g(U)$. In addition, $N(0, 1)$ is a normal distribution and $(v_1, v_2) = (2, 0.84)$ denotes the coefficients that change σ based on the result of the mutation.

TABLE II
ALGORITHM OF THE STATIC-IRT TEST ASSEMBLY MECHANISM FOR FML

| |
|---|
| **Input:** |
| 1.   σ ←1 |
| 2.   $a_{i_1}, a_{i_2}, …, a_{i_l}, b_{i_1}, b_{i_2}, …, b_{i_l}, c_{i_1}, c_{i_2}, …, c_{i_l}$: Parameters $a$, $b$, $c$ for all of the items $i_1, i_2, …,$ and $i_l$, where $1 \le i \le |I|$ |
| 3.   $U \leftarrow (u_{i_1}, u_{i_2}, …, u_{i_l})$: a vector with random values to show an importance of all of the items $i_1, i_2, …, i_l$ |
| 4.   $P_0$: priori distribution |
| **Output:** |
|     The most suitable item set with $M$ selected items for the specific students, where $1 \le M \le |I|$ |
| **Method:** |

**Step1:** Sort $U$ to make $u_{i1} \geq u_{i2} \geq \ldots \geq u_{il}$, where $i_1$, $i_2$, …, and $i_l$ are the indices such that $u_{i1} \geq u_{i2} \geq \ldots \geq u_{il}$
**Step2:** Select the first $M$ items, that is, $i_1$, $i_2$, …, and $i_M$ from the sorted $U$
  **Step2.1:** Simulate $S$ students with different abilities, that is, $\theta_1$, $\theta_2$, …, and $\theta_S$ to make a response to the selected items $i_1$, $i_2$, …, and $i_M$
  **Step2.2:** Simulate $S$ students' response data $r_1$, $r_2$, …, and $r_M$ to items $i_1$, $i_2$, …, and $i_M$ by randomizing a value $r$ on the interval of [0, 1]

    **Step2.2.1:** $r_{sm} \leftarrow 1$ if $r \leq c_{i_m} + (1-c_{i_m}) \times \dfrac{1}{1+e^{-1.7 \times a_{i_m}(\theta_s - b_{i_m})}}$ , where

$1 \leq s \leq |S|$ and $1 \leq m \leq M$
    **Step2.2.2:** $r_{sm} \leftarrow 0$
  **Step2.3:** Execute IRT-based Bayesian estimation mechanism to estimate the ability for each simulated student to get $\widehat{\theta}_s$ by inputting a response pattern matrix $R(s, m)$ with $S$ rows and $M$ columns where $1 \leq s \leq S$, $1 \leq m \leq M$, and each element of $R(s, m)$ is $r_{sm}$
  **Step2.4:** Calculate $g(U) \leftarrow \sum_{s=1, 2, \ldots, S} P_0(\theta_s) \times (\theta_s - \widehat{\theta}_s)^2$
**Step3:** End

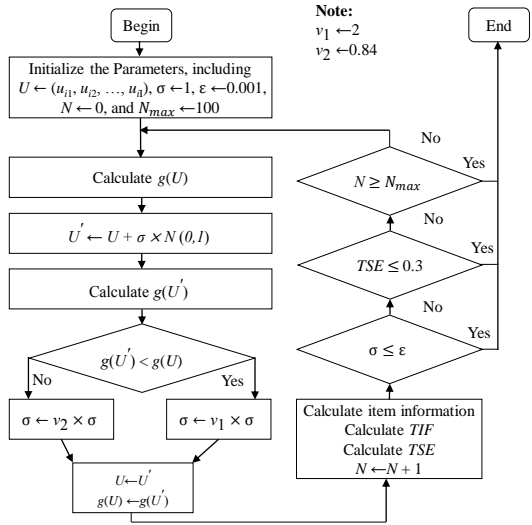

Fig. 3. Flowchart for optimizing the parameters of objective function.

### D. Gauss-Seidel (GS)-based parameter estimation mechanism for Fuzzy Markup Language

Assume that there is a response pattern matrix $R(S, I)$ for $|S|$ students and $|I|$ items, and its size is $|S| \times |I|$. Each element is either 1 (correct response to the item) or 0 (incorrect response to the item). The GS-based parameter estimation mechanism is composed of three stages and repeats these three stages until convergence is achieved or the termination conditions are satisfied. Fig. 4 shows the flowchart of the proposed GS-based parameter estimation mechanism.

1) **Stage 1**: We estimate $|S|$ students' abilities by assigning the default parameters of $a = 1$, $b = 0$, and $c = 0.25$ to $|I|$ items, where the default parameter values are obtained according to the following reasons.

- The center of the item bank is $a = 1$, $b = 0$, and $c = 0.25$. The IRT parameters of the item bank were estimated according to the total of the items' difficulty $b$ (with mean = 0 and standard deviation = 1), discrimination $a = 1$, as well as guessing $c = \dfrac{1}{\text{number of options}} = 0.25$, where a multiple-choice item has four options (one correct answer and three distractors). The relationship to the students' abilities is as follows: According to the IRT, the scale of the estimation of the difficulty of the item and a student's ability are the same; therefore, we can use his / her response data to match the subsequent item's difficulty.

- We use the center of the item bank as the starting point to select the next suitable items for the students based on their response data. When a student chooses an incorrect response, an easier item ($b < 0$) is given. By contrast, when a student chooses correctly, a more difficult item ($b > 0$) is given.

- We obtain ability $\widehat{\theta}_{|S|}$ using (5) [20, 22].

$$f(\mathbf{x}) = f(\boldsymbol{\theta}, \mathbf{a}, \mathbf{b}, \mathbf{c}|\mathbf{u}) \propto L(\mathbf{u}|\boldsymbol{\theta}, \mathbf{a}, \mathbf{b}, \mathbf{c}) \left[ \prod_{i=1}^{|I|} f(a_i) f(b_i) f(c_i) \right] \left[ \prod_{s=1}^{|S|} f(\theta_s) \right] \quad (5)$$

where

- $\boldsymbol{\theta} = \{\theta_1, \theta_2, \ldots, \theta_{|S|}\}$ denotes $|S|$ students' ability set,

- $\mathbf{a} = \{a_1, a_2, \ldots, a_{|I|}\}$, $\mathbf{b} = \{b_1, b_2, \ldots, b_{|I|}\}$, and $\mathbf{c} = \{c_1, c_2, \ldots, c_{|I|}\}$ denote items' parameter sets for item set $I$, respectively, and $\mathbf{x} = \{\boldsymbol{\theta}, \mathbf{a}, \mathbf{b}, \mathbf{c}\}$,

- $\mathbf{u} = \{u_{11}, u_{12}, \ldots, u_{|S||I|}\}$ denotes the response data matrix with $|S|$ students and $|I|$ items,

- $f(\boldsymbol{\theta}, \mathbf{a}, \mathbf{b}, \mathbf{c}|\mathbf{u})$ is the joint posterior density function,

- $L(\mathbf{u}|\boldsymbol{\theta}, \mathbf{a}, \mathbf{b}, \mathbf{c})$ is a likelihood function for students' abilities and item parameters based on $\mathbf{u}$,

- $f(\theta_i)$, $f(a_i)$, $f(b_i)$, and $f(c_i)$ are the prior distributions for $\theta_s$, $a_i$, $b_i$, and $c_i$, respectively, where $1 \leq s \leq |S|$ and $1 \leq i \leq |I|$,

- $f(\theta_i)$ and $f(b_i)$ are a normal distribution function, $f(a_i)$ is a chi distribution function, and $f(c_i)$ is a beta distribution function [20, 22].

2) **Stage 2**: Standardize the estimated abilities using (6) [20].

$$z = \dfrac{x - \bar{\mu}}{\sigma} \quad (6)$$

where, $z$, $x$, $\bar{\mu}$, and $\sigma$ denote the $z$-score, raw data, mean of the raw data, and standard deviation of the raw data, respectively.

3) **Stage 3**: Parameters $b_i$, $a_i$, and $c_i$ for $I$ items are estimated using the acquired $\widehat{\theta}_s$ for $S$ students. The acquired parameters are $\widehat{b}_i$, $\widehat{a}_i$, and $\widehat{c}_i$ ($1 \leq i \leq |I|$).

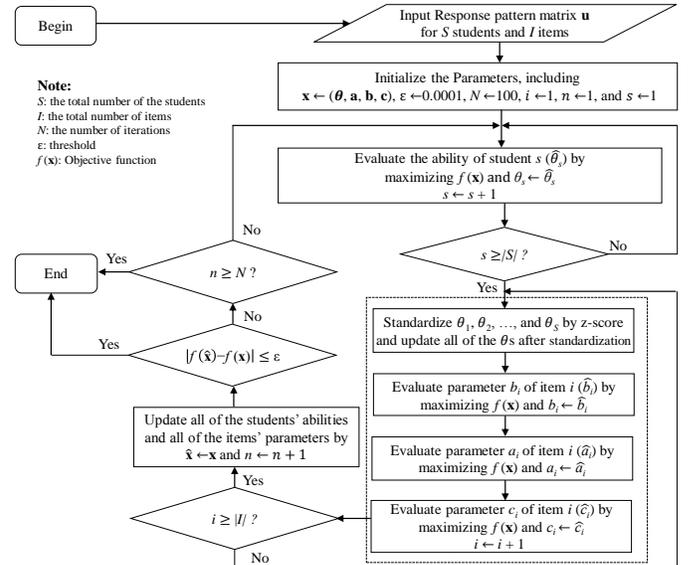

Fig. 4. Flowchart of the proposed GS-based parameter estimation mechanism.

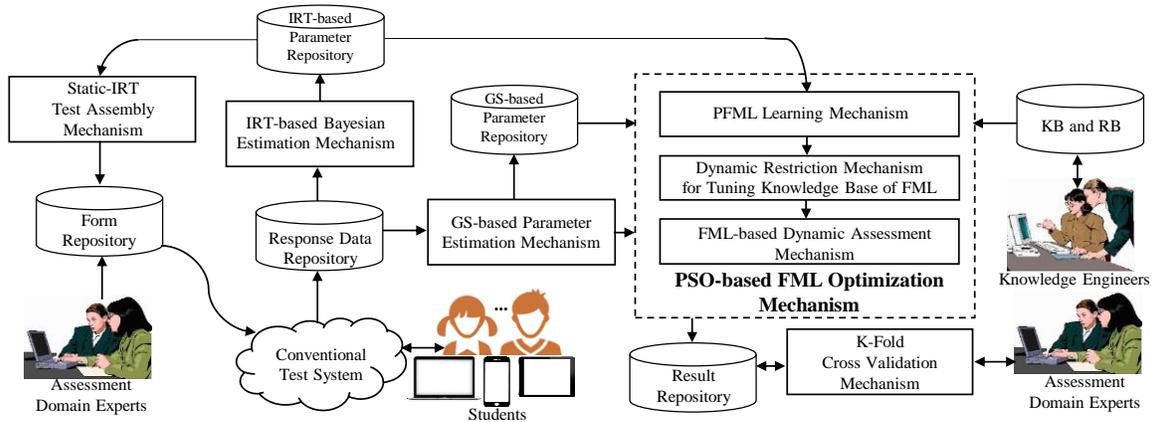

Fig. 5. Structure of PSO-based FML optimization for a dynamic assessment in educational applications.

## III. PSO-BASED FML OPTIMIZATION FOR DYNAMIC ASSESSMENT ON EDUCATIONAL APPLICATION

In this section, we present a PFML optimization for dynamic assessment of students' learning performance and educational applications. First, the proposed structure is introduced. Then, we present an FML-based dynamic assessment mechanism, a PFML learning mechanism, and a dynamic restriction mechanism for tuning the knowledge base of FML.

### A. Proposed Structure

Fig. 5 shows the structure used to integrate the IRT with PFML optimization for a dynamic assessment in educational applications. Fig. 6 shows the entire flowchart of the proposed method. We briefly describe the operation of the proposed structure as follows:

1) A conventional test comprises a pilot test and a norm test. The abilities of the involved student set $S = \{S_1, S_2, \ldots, S_{|S|}\}$ are distributed over a range of ability levels from $-4$ to $+4$ for a conventional test [20]. According to [32], after domain experts finish editing item set $I = \{I_1, I_2, \ldots, I_{|I|}\}$ and assembling the test paper, a pilot test starts. Based on the results of the pilot test, the domain experts modify the item set $I$ and reassemble the test paper. Next, a norm test starts and the students' response data are stored in the response data repository.

2) The IRTBE mechanism iterates to estimate item parameter and ability values based on the students' response data until reaching convergence and stores the estimated values in the IRT-based parameter repository [20, 22].

3) The static-IRT test assembly mechanism assembles the test paper based on the items stored in the IRT-based parameter repository for a given ability scale. The assembled test papers are stored in the form repository after validation by the domain experts.

4) According to the data of $|S|$ students' responses to $|I|$ items, the GS-based parameter estimation mechanism iterates to estimate item parameters until reaching convergence. The estimated results are stored in the GS-based parameter repository. In this study, $|S| = 732$ and $|I| = 51$ were assigned as the input of the PFML optimization mechanism.

5) Knowledge engineers use FML to describe the before-

learning knowledge base and the rule base of the PFML optimization mechanism. The PFML learning mechanism optimizes the inferred results from the FML-based dynamic assessment mechanism reaching until convergence or a maximum generation condition, thus acquiring the after-learning knowledge base and rule base. During optimization, the dynamic restriction mechanism for tuning the knowledge base of FML optimizes the parameters of the fuzzy sets based on the desired output. Finally, we use the K-fold cross validation mechanism to compare the results of the proposed method with those of the IRT-based 3PL model provided by domain experts and store the results in the result repository.

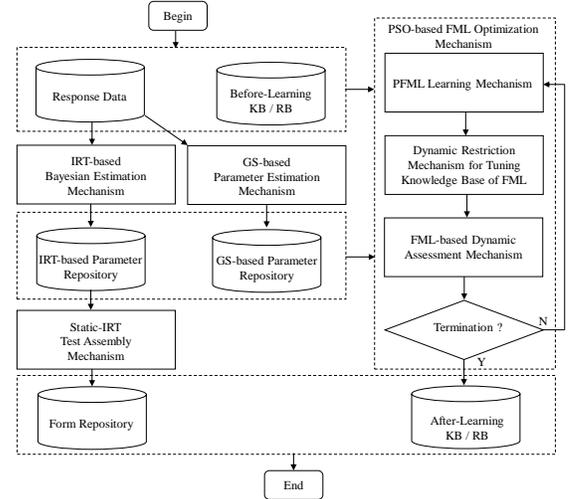

Fig. 6. Entire flowchart of the proposed method.

### B. FML-based Dynamic Assessment Mechanism

The $a$, $b$, and $c$ parameters of an item, which represent its discrimination, difficulty, and guessing, and a student's ability are chosen as the input fuzzy variables *Discrimination*, *Difficulty*, *Guessing*, and *Ability* for the FML-based dynamic assessment mechanism. We define the possibility of correctly answering the item as the output fuzzy variable *Correct_Response_Possibility* (*CRP*). Next, we briefly describe how to construct the knowledge base and rule base of the FML-based dynamic assessment mechanism [29].

1) **Input fuzzy variables *Discrimination*, *Difficulty*, and *Guessing*:** We first use fourth-grade math items to compute

the minimum, maximum, and average values for parameters $a$, $b$, and $c$, respectively. Second, we consider the following standard deviations: 1) $STD_1$: the standard deviation for items whose parameters are between the minimum and average, 2) $STD_2$: the standard deviation for items whose parameters are between the average and maximum, and 3) $STD_3$: the standard deviation for all items. On these basis of these values, we construct the knowledge base of the fuzzy variables *Discrimination*, *Difficulty*, and *Guessing*. Reference [29] shows the values of minimum, maximum, average, $STD_1$, $STD_2$, and $STD_3$ of parameters $a$, $b$, and $c$ for the fourth-grade math items.

2) **Input fuzzy variable *Ability*:** The National Assessment of Educational Progress in the United States [33] uses four levels to define student performance, namely below basic, basic, proficient, and advanced. Table III shows the corresponding ranges of T score ($T\_score$) and *Ability* ($\theta$) for the different student performance levels [19]. The relationship between $T\_score$ and $\theta$ is computed using (7) [19, 20]. Therefore, we define *Ability* to have four fuzzy sets corresponding to below basic, basic, proficient, and advanced.

$$T\_score = \theta \times 10 + 50 \tag{7}$$

### TABLE III
T SCORE AND ABILITY RANGES FOR FOUR LEVELS

| Level | $T\_score$ | $\theta$ |
|---|---|---|
| *Below Basic* | $T\_score < 40$ | $\theta < -1$ |
| *Basic* | $40 \leq T\_score < 54$ | $-1 \leq \theta < -0.4$ |
| *Proficient* | $54 \leq T\_score < 65$ | $-0.4 \leq \theta < 1.5$ |
| *Advanced* | $65 \leq T\_score$ | $1.5 \leq \theta$ |

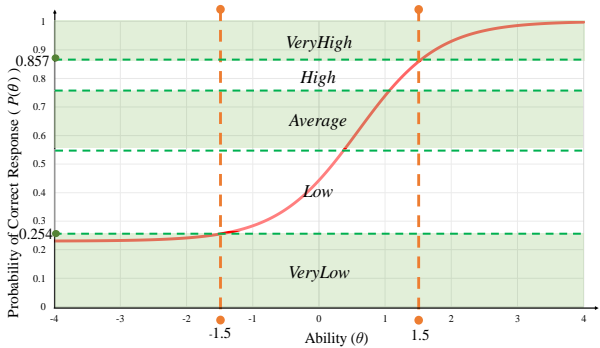

Fig. 7. ICC when $a = 0.96$, $b = 0.59$, and $c = 0.23$ [29].

3) **Output fuzzy variable *Correct_Response_Possibility*:** The parameters of *CRP* are constructed based on an ICC whose parameters $a$, $b$, and $c$ are the averages of the parameters for all fourth-grade math items. Fig. 7 shows that the curve between $\theta = -1.5$ and $\theta = 1.5$ is the steepest and that $P(-1.5) = 0.254$ and $P(1.5) = 0.857$. Hence, we divide probability of correct response into three intervals: 1) between 0.254 and 0.857 for fuzzy sets with the linguistic terms *Low*, *Average*, and *High*; 2) lower than 0.254 for a fuzzy set with the linguistic term *VeryLow*; and 3) higher than 0.857 for a fuzzy set with the linguistic term *VeryHigh*. We utilize the trapezoidal membership function shown as (8) for fuzzy set *FS*, specified by four parameters expressed as *BS*, *BC*, *EC*, and *ES*. Table IV shows the parameters of the defined fuzzy sets.

$$FS\,(x\colon BS,\ BC,\ EC,\ ES) = \begin{cases} 0, & x \leq BS \\ \dfrac{x\text{-}BS}{BC\text{-}BS}, & BS \leq BC \\ 1, & BC \leq x \leq EC \\ \dfrac{ES\text{-}x}{ES\text{-}EC}, & EC < x \leq ES \\ 0, & x > ES \end{cases} \tag{8}$$

where *BS*, *BC*, *EC*, and *ES* denote the *Begin Support*, *Begin Core*, *End Core*, and *End Support*, respectively, of the trapezoidal-shape fuzzy set *FS*.

### TABLE IV
PARAMETERS OF FUZZY SETS

| Discrimination | | Guessing | |
|---|---|---|---|
| **Fuzzy Set** | **[BS, BC, EC, ES]** | **Fuzzy Set** | **[BS, BC, EC, ES]** |
| Low | [0, 0, 0.65, 0.74] | Low | [0, 0, 0.17, 0.19] |
| Medium | [0.67, 0.82, 1.11, 1.25] | Medium | [0.18, 0.21, 0.26, 0.28] |
| High | [1.17, 1.42, 2, 2] | High | [0.26, 0.33, 1, 1] |

| Difficulty | | Ability | |
|---|---|---|---|
| **Fuzzy Set** | **[BS, BC, EC, ES]** | **Fuzzy Set** | **[BS, BC, EC, ES]** |
| VeryEasy | [-4, -4, -1.1, -0.6] | BelowBasic | [-4, -4, -1.1, -0.6] |
| Easy | [-1.0, -0.65, 0.05, 0.4] | Basic | [-1.0, -0.65, 0.05, 0.4] |
| Average | [0.05, 0.4, 0.95, 1.5] | Proficient | [0.05, 0.4, 0.95, 1.5] |
| Hard | [0.95, 1.5, 4, 4] | Advanced | [0.95, 1.5, 4, 4] |

| Correct_Response_Possibility (CRP) | | | |
|---|---|---|---|
| **Fuzzy Set** | **[BS, BC, EC, ES]** | **Fuzzy Set** | **[BS, BC, EC, ES]** |
| VeryLow | [0, 0, 0.23, 0.34] | High | [0.58, 0.8, 0.8, 0.97] |
| Low | [0.23, 0.34, 0.34, 0.58] | VeryHigh | [0.8, 0.96, 1, 1] |
| Average | [0.34, 0.58, 0.58, 0.80] | | |

4) **Rule base:** Table V briefly describes the method for constructing the rule base of the FML-based dynamic assessment mechanism, and Table VI shows part of the adopted FML.

### TABLE V
ALGORITHM FOR CONSTRUCTING FUZZY RULE OF FML

**Input:** Fuzzy variables and fuzzy sets of FML
**Output:** Fuzzy rule of FML
**Method:**
  **Step1:** $a\_linguisticterm \leftarrow$ fuzzy set of fuzzy variable *Discrimination* from knowledge base of FML
  **Step2:** $a \leftarrow Discrimination\_linguisticterm\_BC$
  **Step3:** $b\_linguisticterm \leftarrow$ fuzzy set of fuzzy variable *Difficulty* from knowledge base of FML
  **Step4:** $b \leftarrow Difficulty\_linguisticterm\_BC$
  **Step5:** $c\_linguisticterm \leftarrow$ fuzzy set of fuzzy variable *Guessing* from knowledge base of FML
  **Step6:** $c \leftarrow Guessing\_linguisticterm\_BC$
  **Step7:** $\theta\_linguisticterm \leftarrow$ fuzzy set of fuzzy variable *Ability* from knowledge base of FML
  **Step8:** $\theta \leftarrow Ability\_linguisticterm\_BC$
  **Step 9:** Calculate probability of correct response and assign output to $p$
  **Step 10:** Calculate membership degree and find the maximum
    **Step 10.1:** $\mu_s \leftarrow FS$
    **Step 10.2:** $\mu_{max} = Max(\mu_s)$
    **Step 10.3:** $FS = \{CRP\_VeryLow,\ CRP\_Low,\ CRP\_Average,\ CRP\_High,\ CRP\_VeryHigh\}$
  **Step 11:** $p\_linguisticterm \leftarrow$ linguistic term with the maximum membership degree ($\mu_{max}$) when the input of fuzzy set is $p$
  **Step12:** Compose the fuzzy rule
    **Step12.1:** If *Discrimination* is $a\_linguisticterm$ and *Difficulty* is $b\_linguisticterm$ and *Guessing* is $c\_linguisticterm$ and *Ability* is $\theta\_linguisticterm$ Then *CRP* is $p\_linguisticterm$
  **Step 13:** End.

### TABLE VI
PARTIAL OF ADOPTED FML

```xml
<?xml version="1.0"?>
<FuzzyController ip="localhost" name="">
    <KnowledgeBase>
        <FuzzyVariable domainleft="0" domainright="2" name="Discrimination" scale="" type="input">
            <FuzzyTerm name="Low" hedge="Normal">
```

```
            <TrapezoidShape Param1="0" Param2="0" Param3="0.65" Param4="0.74" />
          </FuzzyTerm>
          <FuzzyTerm name="Medium" hedge="Normal">
            <TrapezoidShape  Param1="0.67"  Param2="0.815"  Param3="1.105"
Param4="1.25" />
          </FuzzyTerm>
          <FuzzyTerm name="High" hedge="Normal">
            <TrapezoidShape Param1="1.17" Param2="1.42" Param3="2" Param4="2" />
          </FuzzyTerm>
        </FuzzyVariable>
                          ⋮
      <FuzzyVariable            domainleft="0"            domainright="1"
name="CorrectResponsePossibility" scale="" type="output">
        <FuzzyTerm name="VeryLow" hedge="Normal">
          <TrapezoidShape Param1="0" Param2="0" Param3="0.23" Param4="0.34" />
        </FuzzyTerm>
        <FuzzyTerm name="Low" hedge="Normal">
          <TriangularShape Param1="0.23" Param2="0.34" Param3="0.58" />
        </FuzzyTerm>
        <FuzzyTerm name="Average" hedge="Normal">
          <TriangularShape Param1="0.34" Param2="0.58" Param3="0.8" />
        </FuzzyTerm>
        <FuzzyTerm name="High" hedge="Normal">
          <TriangularShape Param1="0.58" Param2="0.8" Param3="0.97" />
        </FuzzyTerm>
        <FuzzyTerm name="VeryHigh" hedge="Normal">
          <TrapezoidShape Param1="0.8" Param2="0.97" Param3="1" Param4="1" />
        </FuzzyTerm>
      </FuzzyVariable>
    </KnowledgeBase>
    <RuleBase activationMethod="MIN" andMethod="MIN" orMethod="MAX"
name="RuleBase1" type="mamdani">
```

```
<Rule name="Rule1" connector="and"          <Rule            name="Rule144"
weight="1" operator="MIN">                   connector="and"       weight="1"
  <Antecedent>                               operator="MIN">
    <Clause>                                   <Antecedent>
      <Variable>Discrimination</Variable>       <Clause>
      <Term>Low</Term>                            <Variable>Discrimination</Variable>
    </Clause>                                     <Term>High</Term>
    <Clause>                                    </Clause>
      <Variable>Difficulty</Variable>           <Clause>
      <Term>VeryEasy</Term>                       <Variable>Difficulty</Variable>
    </Clause>                                     <Term>Hard</Term>
    <Clause>                                    </Clause>
      <Variable>Guessing</Variable>             <Clause>
      <Term>Low</Term>                            <Variable>Guessing</Variable>
    </Clause>                                     <Term>High</Term>
    <Clause>                                    </Clause>
      <Variable>Ability</Variable>              <Clause>
      <Term>BelowBasic</Term>                     <Variable>Ability</Variable>
    </Clause>                                     <Term>Advanced</Term>
  </Antecedent>                                 </Clause>
  <Consequent>                               </Antecedent>
    <Clause>                                 <Consequent>
                                               <Clause>
<Variable>CorrectResponsePossibility
</Variable>                                  <Variable>CorrectResponsePossibility
      <Term>Average</Term>                   </Variable>
    </Clause>                                    <Term>Average</Term>
  </Consequent>                                </Clause>
</Rule>                                       </Consequent>
                                             </Rule>
              ⋮

    </RuleBase>
  </FuzzyController>
```

## C. PFML Learning Mechanism

This section introduces the PFML learning mechanism by combining FML with PSO. Table VII describes the basic concept of PSO [14, 34]. Each candidate solution is called a particle and represents a point in a $D$-dimensional space, where $D$ denotes the number of parameters to be optimized. After each iteration, the particle's velocity and position are updated until reaching convergence. The relationship between the particle in PSO and parameters of the fuzzy sets is illustrated as follows:

1) One particle is composed of the knowledge base of the FML-based dynamic assessment mechanism. That is, each particle represents the fuzzy-set parameters of the four input fuzzy variables (*Discrimination*, *Difficulty*, *Guessing*, and *Ability*) and the output fuzzy variable (*CRP*). Fig. 8 shows the composed parameters of the $i$th particle ($\mathbf{x}_i = [x_{i1}, x_{i2}, x_{i3}, \ldots x_{i5}]$ in five-dimensional space), in this paper. We utilize the trapezoidal membership function to express the parameters of each fuzzy set. Fig 8 indicates that 1) $x_{i1\text{-}1}$, $x_{i1\text{-}2}$, $x_{i1\text{-}3}$, and $x_{i1\text{-}4}$ denote the parameters $BS$, $BC$, $EC$, and $ES$, respectively, of the fuzzy set *Discrimination_Low* and 2) $x_{i5\text{-}17}$, $x_{i5\text{-}18}$, $x_{i5\text{-}19}$, and $x_{i5\text{-}20}$ denote the parameters $BS$, $BC$, $EC$, and $ES$, respectively, of the fuzzy set *CRP_VeryHigh*. Hence, in this study, the total number of parameters for the $i^{\text{th}}$ particle ($\mathbf{x}_i$) is 76.

TABLE VII
BASIC CONCEPT OF PSO [14, 34]

/*Update the velocity of the $i$th particle*/
$$\mathbf{v}_i(t+1) = w \times \mathbf{v}_i(t) + c_1 \times (\mathbf{pBest}_i - \mathbf{x}_i(t)) \times r_1 + c_2 \times (\mathbf{gBest} - \mathbf{x}_i(t)) \times r_2$$
/*Update the position of the $i$th particle*/
$$\mathbf{x}_i(t+1) = \mathbf{x}_i(t) + \mathbf{v}_i(t+1)$$
where
1) $\mathbf{x}_i = [x_{i1}, x_{i2}, x_{i3}, \ldots x_{iD}]$ /*position of the $i$th particle with $D$ parameters to be optimized*/
2) $\mathbf{X} = \{ \mathbf{x}_1, \mathbf{x}_2, \mathbf{x}_3, \ldots, \mathbf{x}_N \}$ /*population of $N$ candidate solutions constitutes the swarm*/
3) $i$ is particle index, $t$ is the iteration index, and $w$ is inertia weight
4) $c_1$ (cognitive parameter) and $c_2$ (social parameter) are acceleration constants, $\mathbf{pBest}_i$ is the best position of the $i^{\text{th}}$ particle where $\mathbf{pBest}_i = [pBest_{i1}, pBest_{i2}, pBest_{i3}, \ldots, pBest_{iD}]$, as well as $\mathbf{gBest}$ is the best position among all $N$ particles in the swarm where $\mathbf{gBest} = [gBest_1, gBest_2, gBest_3, \ldots, gBest_D]$
5) $r_1$ and $r_2$ are the random values generated from a uniform distribution in [0, 1]

2) In this study, a swarm is composed of 20 particles ($N = 20$). The parameters of four input fuzzy variables and one output fuzzy variable represent the position of the particle in five-dimensional space; they are optimized by adjusting the moving velocity in order to reach convergence.

3) The domain of the particle in each dimension is contained in [*domainleft*, *domainright*] of the fuzzy variable. In this paper, the domains of the first to the fifth dimensions are [0, 2], [−4, 4], [0, 1], [−4, 4], and [0, 1]. These domains are employed in the optimization of the parameters of *Discrimination*, *Difficulty*, *Guessing*, *Ability*, and *CRP*, respectively.

4) Fig. 9 shows the flowchart of the PFML learning mechanism. In this study, we assigned $w = 0$, $c_1 = 2$, and $c_2 = 2$. In addition, the PFML learning mechanism terminates when the iteration number is greater than 1000 or when fitness is less than 0.001. The fitness function *Fitness*($x_i, y_i$) is calculated using (9). After termination, we use the positions stored in $\mathbf{gBest}$ to compose the after-learning knowledge base of the FML-based dynamic assessment mechanism. Section III.D introduces the details of the dynamic restriction mechanism for tuning the knowledge base of FML.

$$Fitness(x_i, y_i) = \sum_{i=1}^{M}(x_i - y_i)^2 / M \qquad (9)$$

where $M$ denotes the total number of the data points as well as $x_i$ and $y_i$ denote the inferred result and desired output of the $i^{\text{th}}$ data point, respectively.

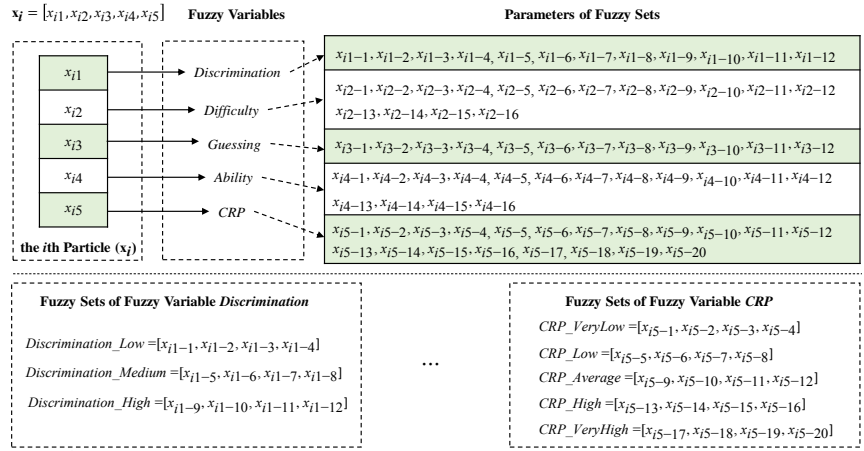

Fig. 8. Composed parameters of the $i^{th}$ particle ($\mathbf{x}_i$).

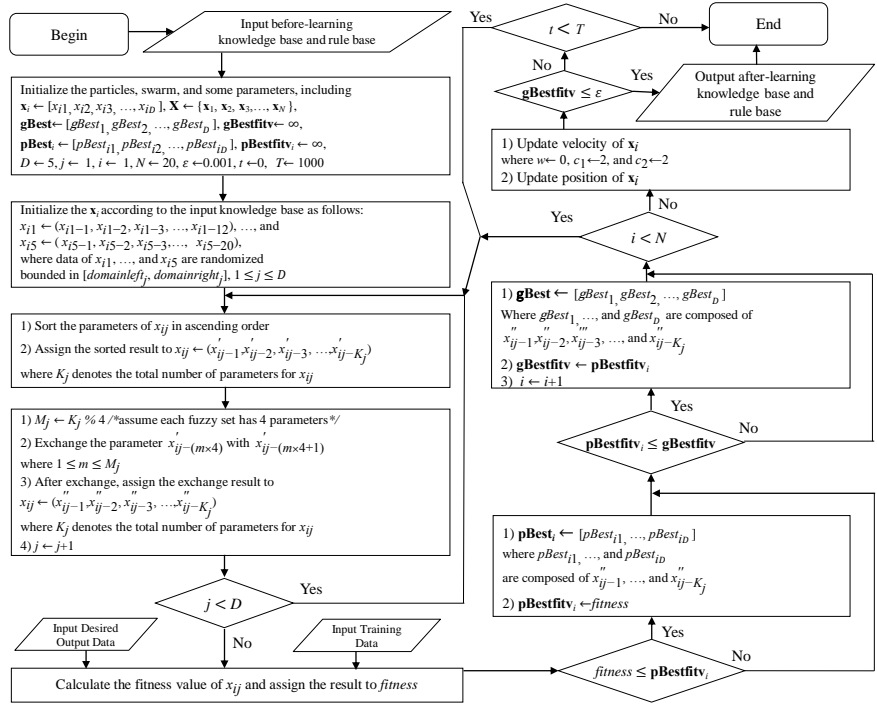

Fig. 9. Flowchart of PFML learning mechanism.

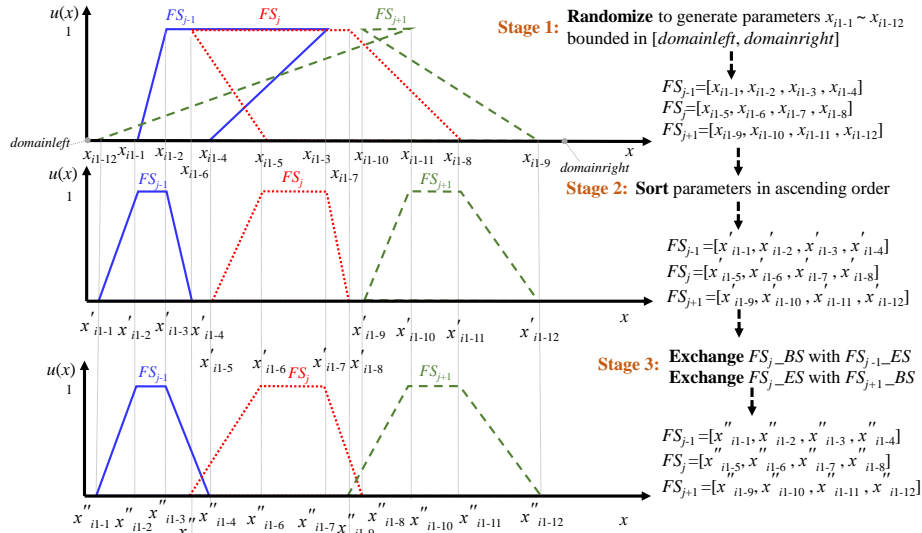

Fig. 10. Three-stage dynamic restriction mechanism for tuning the knowledge base of FML.

### D. Dynamic Restriction Mechanism for Tuning Knowledge Base of FML

This section introduces the proposed dynamic restriction mechanism for tuning the knowledge base of FML, which includes the randomization, sort, and exchange stages. Assume there is a fuzzy variable $x$ with three fuzzy sets, $FS_{j-1}$, $FS_j$, and $FS_{j+1}$. Each fuzzy set has four parameters so the $i^{th}$ particle in the first dimension has 12 parameters, $x_{i1-1}$, $x_{i1-2}$, ..., and $x_{i1-12}$. Fig. 10 illustrates the three-stage dynamic restriction mechanism for tuning the knowledge base of FML. We briefly describe the operation as follows:

1) **Stage 1: Randomize** to generate initial parameters $x_{i1-1}$, $x_{i1-2}$, ..., $x_{i1-12}$ between *domainleft* and *domainright*.

2) **Stage 2: Sort** the parameters in ascending order to acquire the newly generated 12 parameters, namely $x'_{i1-1}$, $x'_{i1-2}$, ..., and $x'_{i1-12}$.

3) **Stage 3: Exchange** the values of *BS* and *ES* for two adjacent fuzzy sets to meet the requirements for constructing the parameters of a fuzzy variable. For example, $FS_j\_BS$ is exchanged with $FS_{j-1}\_ES$, and $FS_j\_ES$ is exchanged with $FS_{j+1}\_BS$ to acquire the new 12 parameters, namely $x''_{i1-1}$, $x''_{i1-2}$, ..., and $x''_{i1-12}$, of fuzzy sets $FS_{j-1}$, $FS_j$, and $FS_{j+1}$.

## IV. EXPERIMENTAL RESULTS

In this section, the experimental environment and the adopted cross validation method are first introduced. Then, we present some experimental results, including the performance evaluations for the static-IRT testing assembly mechanism, GS-based parameter estimation mechanism, FML-based dynamic assessment mechanism, and PFML learning mechanism.

### A. Types of Graphics Introduction to Experimental Environment and Adopted Cross Validation Method

To evaluate the performance of the proposed approach, we used the collected response data from a conventional test held in Kaohsiung, Taiwan, in May and June 2014. There were 32 students from grades two–nine involved in pilot tests on Chinese and math. In addition, 192 students were involved the norm test, and each subject had two types of test papers, which were given to the students. We used the students' response to the fourth-grade-math items as the input data and conducted some experiments to assess the performance. The number of students involved in the fourth-grade math norm test was 732. For the fourth-grade-math test, each test paper included 28 items, five of which were common items; therefore, the total number of items for fourth-grade math was 51.

Fig. 11 shows the adopted K-fold cross validation method that operates as follows. 1) $K = 5$ means that 80% of the students' response data is assigned for training and 20% is for testing. 2) After executing the GS-based parameter estimation mechanism and IRTBE mechanism, the estimated abilities and item parameters are stored in the GS-based parameter repository and IRT-based parameter repository, respectively. 3) We use data from the GS-based parameter repository and the IRT-based parameter repository as the training and test data and the desired output data, respectively. The training data are for the learning data ($D_{Learning}$) and the testing data are for the validating data ($D_{Validating}$). 4) We compare the performances of the proposed and traditional approaches and store the results in the result repository.

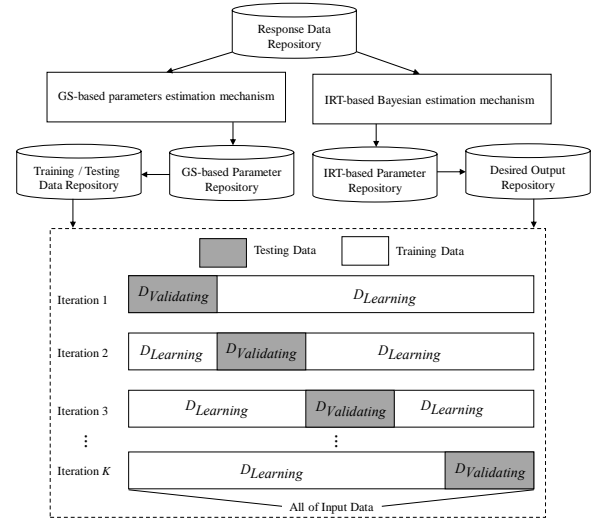

Fig. 11. K-Fold cross validation method.

### B. Performance Evaluation for Static-IRT Test Assembly Mechanism

The first experiment evaluates the performance of the assembly test paper. We assemble the test paper according to four performance levels: below basic, basic, proficient, and advanced. Fig. 12 (a) shows the TIF curves for the four levels and Fig. 12 (b) shows the TIF and TSE curves for the proficient-level students. Figs. 12 (a)–(b) indicate that the peak ability is relatively higher when the test paper is assembled for the higher-level students and that the peak of the TIF curve matches the minimum value of the TSE.

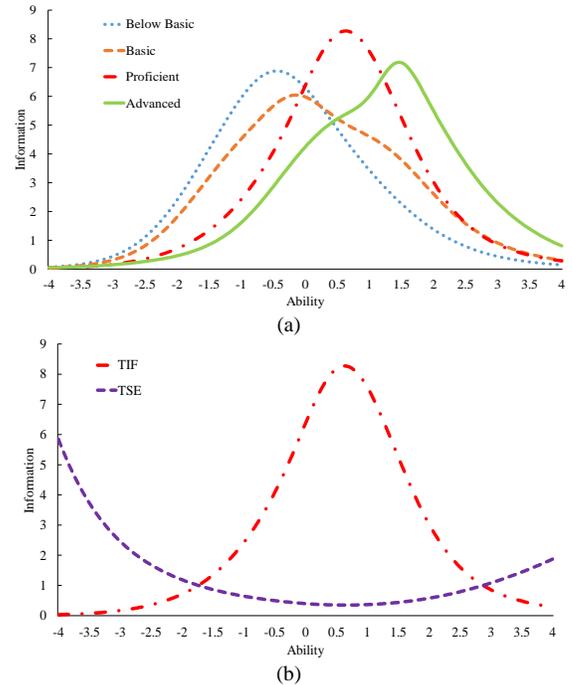

Fig. 12. (a) *TIF* curves for *below basic*, *basic*, *proficient*, and *advanced* students and (b) *TIF* and *TSE* curves for *proficient* students.

TABLE VIII
RESPONSE DATA FROM $S$=732 STUDENTS AND $I$=51 ITEMS

| Item No. / Student No. | 1 | 2 | 3 | 4 | 5 | 6 | 7 | 8 | 9 | 10 | | 42 | 43 | 44 | 45 | 46 | 47 | 48 | 49 | 50 | 51 |
|---|---|---|---|---|---|---|---|---|---|---|---|---|---|---|---|---|---|---|---|---|---|
| 1 | 1 | 1 | 1 | 1 | 1 | 1 | 0 | 1 | 1 | 1 | | N | N | N | N | N | N | N | N | N | N |
| 2 | 1 | 0 | 1 | 1 | 0 | 0 | 1 | 1 | 1 | 1 | ... | N | N | N | N | N | N | N | N | N | N |
| 3 | 1 | 1 | 1 | 0 | 1 | 1 | 0 | 0 | 1 | 1 | | N | N | N | N | N | N | N | N | N | N |
| 4 | 1 | 1 | 1 | 0 | 1 | 1 | 0 | 0 | 0 | 1 | | N | N | N | N | N | N | N | N | N | N |
| 5 | 1 | 1 | 1 | 0 | 1 | 0 | 1 | 1 | 1 | 1 | | N | N | N | N | N | N | N | N | N | N |
| | | | | | | | | | | | ⋮ | | | | | | | | | | |
| 728 | N | N | N | N | N | N | N | N | N | N | | 1 | 1 | 1 | 1 | 1 | 1 | 1 | 1 | 1 | 1 |
| 729 | N | N | N | N | N | N | N | N | N | N | | 1 | 1 | 0 | 0 | 0 | 1 | 0 | 0 | 1 | 1 |
| 730 | N | N | N | N | N | N | N | N | N | N | ... | 1 | 1 | 0 | 0 | 0 | 0 | 0 | 0 | 1 | 1 |
| 731 | N | N | N | N | N | N | N | N | N | N | | 1 | 0 | 1 | 0 | 0 | 1 | 0 | 1 | 1 | 1 |
| 732 | N | N | N | N | N | N | N | N | N | N | | 0 | 1 | 0 | 0 | 1 | 1 | 0 | 1 | 1 | 1 |

TABLE IX
PARTIAL PARAMETERS ESTIMATED BY GS-BASED MECHANISM AND IRT-BASED MECHANISM PROVIDED BY DOMAIN EXPERTS

| PRM / Item No. | $a_{IRT}$ | $a_{GS}$ | $(a_{GS} - a_{RT})^2$ | $b_{IRT}$ | $b_{GS}$ | $(b_{GS} - b_{IRT})^2$ | $c_{IRT}$ | $c_{GS}$ | $(c_{GS} - c_{IRT})^2$ |
|---|---|---|---|---|---|---|---|---|---|
| 1 | 0.8 | 0.708 | 0.008 | 0.03 | -0.222 | 0.063 | 0.21 | 0.272 | 0.004 |
| 2 | 1.36 | 1.065 | 0.087 | 0.43 | 0.156 | 0.075 | 0.17 | 0.107 | 0.004 |
| 3 | 1.14 | 1.16 | 0 | -0.84 | -1.142 | 0.091 | 0.21 | 0.181 | 0.001 |
| 4 | 0.95 | 0.874 | 0.006 | -0.45 | 0.062 | 0.262 | 0.22 | 0.366 | 0.021 |
| 5 | 0.97 | 0.97 | 0 | -0.11 | -0.791 | 0.463 | 0.22 | 0.152 | 0.005 |
| 6 | 0.44 | 0.752 | 0.097 | -1.93 | -0.357 | 2.476 | 0.27 | 0.619 | 0.122 |
| 7 | 1.23 | 1.168 | 0.004 | 1.93 | 1.729 | 0.04 | 0.32 | 0.315 | 0 |
| 8 | 0.91 | 1.045 | 0.018 | 1.27 | 1.122 | 0.022 | 0.28 | 0.285 | 0 |
| 9 | 0.65 | 0.834 | 0.034 | 0.38 | 0.385 | 0 | 0.26 | 0.295 | 0.001 |
| 10 | 0.95 | 1.016 | 0.004 | 0.88 | 0.701 | 0.032 | 0.2 | 0.18 | 0 |
| | | | | | ⋮ | | | | |
| 48 | 0.93 | 1.14 | 0.044 | 3 | 2.34 | 0.435 | 0.23 | 0.222 | 0 |
| 49 | 0.8 | 1 | 0.04 | 0.99 | 0.825 | 0.027 | 0.19 | 0.183 | 0 |
| 50 | 0.89 | 1.059 | 0.028 | 0.26 | 0.199 | 0.004 | 0.21 | 0.2 | 0 |
| 51 | 0.54 | 0.802 | 0.068 | -1.42 | -0.57 | 0.723 | 0.26 | 0.478 | 0.047 |
| MSE | | 0.0377 | | | 0.1583 | | | 0.0086 | |

## C. Performance Evaluation for GS-based Parameter Estimation Mechanism

The second experiment evaluates the performance of the GS-based parameter estimation mechanism. Table VIII shows the partial response data from the 732 students and the 51 items, where 1, 0, and N denote "*correctly answered the item,*" "*incorrectly answered the item,*" and "*missing data,*" respectively. Herein, "*missing data*" indicates that this student did not choose a response to this item. The GS-based parameter estimation mechanism estimates the parameters of the 51 items until reaching convergence. Table VIII presents the response matrix of the 51 items and 732 students. In Table VIII, columns represent items and rows represent students; the cells are the students' response data for the items. The purpose of Table VIII is to estimate all of the students' abilities through by increasing the number of items and linking the common items based on the limited student response data for different and common items. Table IX shows the partial parameters estimated using the proposed method ($a_{GS}$, $b_{GS}$, and $c_{GS}$) and those provided by the domain experts ($a_{IRT}$, $b_{IRT}$, and $c_{IRT}$) based on the IRT 3PL model. The fitness function utilized to evaluate the parameter estimation is the mean square error (MSE) shown in (9), where $M$ denotes the number of items, and $x$ and $y$ denote the parameters estimated by the proposed method and those provided by the domain experts, respectively. Table IX indicates that the MSE values for parameters $a$, $b$, and $c$ are 0.0377, 0.1583, and 0.0086, respectively. This indicates that the proposed method's performance is similar to the estimated results of the traditional IRT-based 3PL model.

## D. Performance Evaluation for FML-based Dynamic Assessment Mechanism

The third experiment evaluates the precision and recall of the proposed FML-based dynamic assessment mechanism. Precision and recall are calculated using (10) and (11), respectively. After removing the missing data from the response pattern matrix, we use 17,052 response data points as the experimental data, of which 13,608 are for training data and 3444 are for testing data. Additionally, we use 5-fold cross validation to evaluate the performance of the proposed method. Table X briefly presents the method of calculating the precision and recall for the training data and testing data. Fig. 13 shows that the proposed method has superior performance to the IRT-based 3PL model in terms of the precision of training data; however, for recall, the proposed method does not surpass the IRT-based 3PL model.

$$Precision = \frac{TP}{TP + FP} \qquad (10)$$

$$Recall = \frac{TP}{TP + FN} \qquad (11)$$

$$True\ Positive\ Rate\ (TPR) = \frac{TP}{TP + FN} \qquad (12)$$

$$False\ Positive\ Rate\ (FPR) = \frac{FP}{FP + TN} \qquad (13)$$

where $TP$ denotes correct positive classifications, $TN$ denotes correct negative classifications, $FP$ denotes incorrect positive classifications, and *FN* denotes incorrect negative classifications [21].

TABLE X
METHOD OF CALCULATING PRECISION AND RECALL FOR THE TRAINING DATA AND TESTING DATA.

- Predicted values of the training data
- Execute the GS-based parameter estimation mechanism to estimate the items' parameters and students' real-time abilities of the training data.
- Execute the FML-based dynamic assessment mechanism to infer the *CRP* values of the training data based on the estimated items parameters and abilities.
- Calculate the probability of correct response for the training data based on IRT-based 3PL model.
- Predicted values of the testing data
- Utilize the IRT-based Bayesian estimation mechanism to estimate the involved students' real-time abilities based on the estimated items' parameters.
- Execute the FML-based dynamic assessment mechanism to infer the *CRP* for the testing data.
- Use the IRT-based 3PL model to calculate probability of correct response for the testing data.
- Actual values of the training data and testing data
- Extract actual values from the involved students' response data to all of the items, where 1 and 0 denote correct response and incorrect response, respectively, to this item.
- Precision and recall computation for the training data and testing data
- Use thresholds, bounded in the interval [0, 1], to classify the predicted values of the training data and testing data into 0 or 1.
- Calculate precision and recall for each threshold using predicted value and actual value.

second PFML learning method has the highest performance in the experiment. In addition, the higher the generation is, the lower the MSE is. Figs. 15 (a)–(e) show the after-learning fuzzy sets for the fuzzy variables of *Discrimination*, *Difficulty*, *Guessing*, *Ability*, and *CRP*, respectively, according to the second proposed PFML learning method.

TABLE XI
OPERATIONS FOR EVALUATING PFML LEARNING MECHANISM.

- Two kinds of PFML learning mechanisms are implemented:
- **PFML Learning Method No. 1:** For a fuzzy variable, after-learning *BS* value of the first fuzzy set can be greater than or equal to its *domainleft* value and after-learning *ES* value of the last fuzzy set can be less than or equal to its *domainright* value.
- **PFML Learning Method No. 2:** For a fuzzy variable, after-learning *BS* value of the first fuzzy set must be equal to its *domainleft* value and after-learning *ES* value of the last fuzzy set must be equal to its *domainright* value.
- Compare its performance with the GFML-based learning method [8–11] under different generations by *MSE*.
- Compare *precision*, *recall*, and *receiver operating characteristic* (*ROC*) curves with various approaches, including 1) **IRT-based 3PL model**, 2) **FML-based dynamic assessment mechanism**, 3) **GFML-based learning method**, 4) **PFML Learning Method No. 1**, and 5) **PFML Learning Method No. 2**. The *x*-axis and *y*-axis of *ROC* denote *True Positive Rate* (*TPR*) and *False Positive Rate* (*FPR*), respectively. *TPR* and *FPR* are calculated using (12) and (13), respectively.

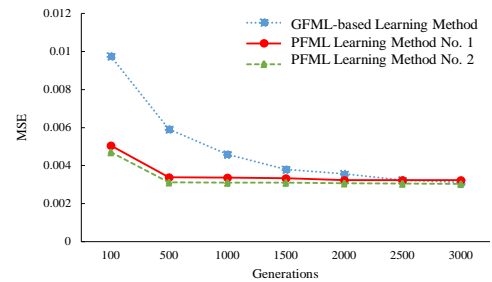

Fig. 14. MSE values under different evolution generations.

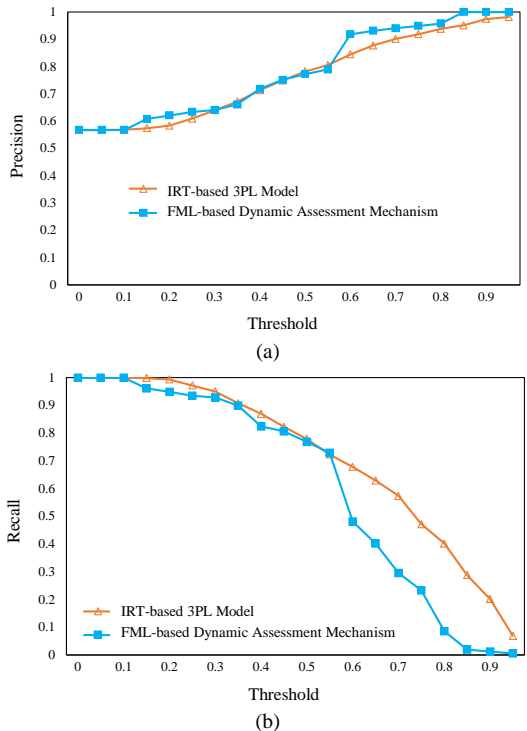

Fig. 13. (a) Precision and (b) recall of the training data.

### E. Performance Evaluation for PFML Learning Mechanism

The fourth experiment shows the performance of the PFML learning mechanism. Table XI lists its operations. Fig. 14 presents the MSE values under different evolution generations by using the genetic FML (GFML)-based learning method [8]–[11] and two PFML learning methods. Fig. 14 indicates that the

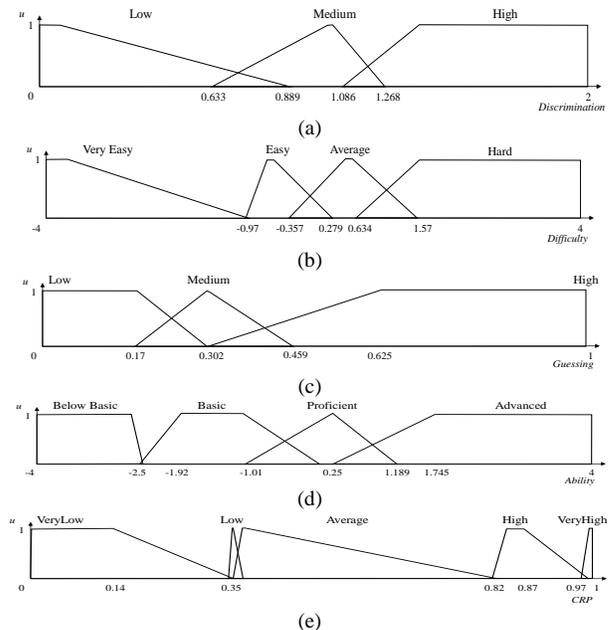

Fig. 15. After-learning fuzzy sets for fuzzy variables (a) *Discrimination*, (b) *Difficulty*, (c) *Guessing*, (d) *Ability*, and (e) *CRP* by PFML learning method No. 2.

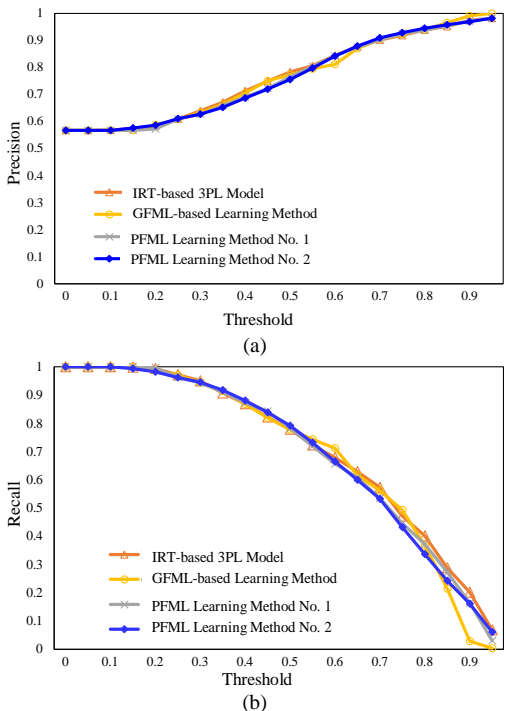

(a)

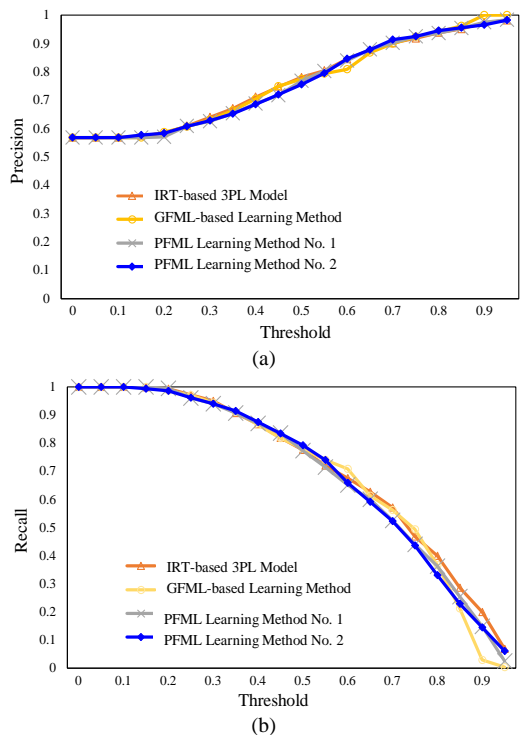

(b)

Fig. 16. (a) Precision and (b) recall of the training data

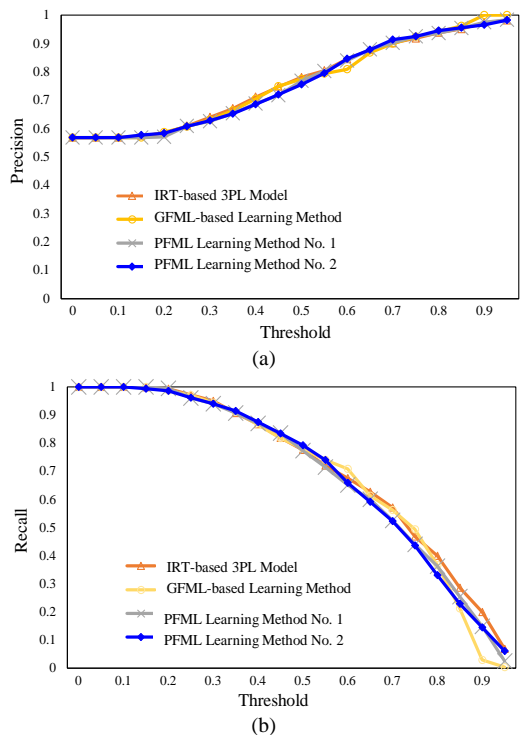

(a)

(b)

Fig. 17. (a) Precision and (b) recall of the testing data.

Figs. 16 (a)–(b) show the precision and recall of the training data, respectively, implemented with the IRT-based 3PL model, the GFML-based learning method, and the two proposed PFML learning methods. Figs. 17(a)–(b) show the precision and recall curves for the testing data. Comparing Fig. 13 with Figs. 16 and 17 reveals that the recall performance is improved after learning. However, there is a large variance in recall performance in the GFML-based learning when the threshold is approximately from 0.8 to 0.95. The recall performance of the PFML learning method, by contrast, is considerably more stable and its trend is more similar to the IRT-based 3PL model than the GFML-based learning method. Figs. 18(a)–(b) show the receiver operating characteristic curve (ROC) curves for the training data and testing data, respectively, implemented with the IRT-based 3PL model, the GFML-based learning method, and the PFML learning methods. Fig. 18 shows that the ROC curve of the IRT-based 3PL model is better than the curve of FML-based dynamic assessment mechanism and that all after-learning ROC curves perform better than the curves of the FML-based dynamic assessment mechanism and IRT-based 3PL model.

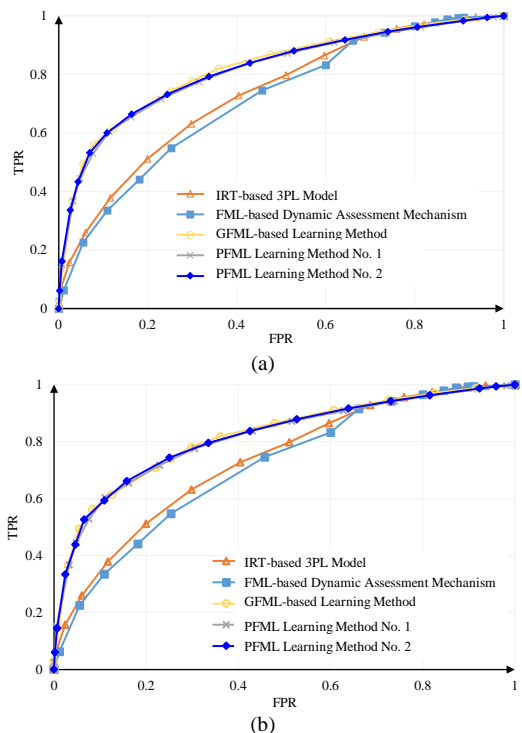

(a)

(b)

Fig. 18. ROC curves for the (a) training data and (b) testing data.

## V. CONCLUSION AND DISCUSSION

This paper proposes a dynamic assessment agent based on FML and PSO for students' learning performance evaluation and educational applications. The core technologies are as follows: 1) The GS-based parameter estimation mechanism estimates items' parameters according to the response data of the conventional test. 2) The static-IRT test assembly mechanism assembles a form for the conventional test. 3) The proposed FML-based dynamic assessment agent infers the probability of correctly answering the item for students with various abilities. 4) The proposed PFML learning mechanism optimizes the knowledge base of the FML to make the experimental results approach to the desired output provided by the domain experts. From the experimental results, the proposed approach performs favorably, especially after executing the PFML learning mechanism. This result indicates that using computational-intelligences technologies is feasible for helping domain experts validate the correctness of education testing statistics in addition to using the popular IRT-based 3PL model. Hence, the dynamic assessment agent is not proposed to replace IRT. It combines PSO machine learning mechanism with IRT and applies it to education domain, especially for future humans and robots co-learning [35, 36]. Additionally, to distinguish items from students' abilities, the proposed method shows the estimated performance level, namely below basic, basic, proficient, and advanced. This method can also concretely describe a student's performance and provide students with their zone of proximal development (ZPD) based on the performance characteristics of the adjacent level. Moreover, the proposed method can save the traditional paper-and-pencil (P&P) test time and provide more accurate estimates of student what can do and what cannot do. Therefore, the proposed agent can help teachers evaluate students learning performance more efficiently, and it is recommended to use in the future classrooms for students and robots co-learning [35, 36] based on IRT, but not to replace item response theory with the proposed method in this paper.

However, some weaknesses exist in the performance of the proposed method; therefore, improvements could be made, for example, by considering the item contents when assembling the test papers, involving the rule base of the FML to be optimized by the PFML learning mechanism, and applying the proposed methods to learning-material recommendations and computer Go learning for students with different performance levels. In the future, we will combine the proposed agent with a robot to build an intelligent robot with co-learning ability, which may provide a key influence on future education and a clear and specific direction for the future of pedagogy.

ACKNOWLEDGMENT

The authors also would like to thank involved students for conventional test, domain experts for their valuable comments on assessment domain knowledge as well as Sheng-Lun Cho and Koun-Hong Lin for their help with organizing the response data of the involved students.

REFERENCES

[1] H. Ishibuchi, "IEEE Standards," *IEEE Computational Intelligence Magazine*, vol. 11, no. 4, pp. 2, Nov. 2016.

[2] G. Acampora, B. D. Stefano, and A. Vitiello, "IEEE 1855TM: The first IEEE standard sponsored by IEEE Computational Intelligence Society," *IEEE Computational Intelligence Magazine*, vol. 11, no. 4, 4-6, Nov. 2016.

[3] G. Acampora and V. Loia, "Fuzzy control interoperability and scalability for adaptive domotic framework," *IEEE Transactions on Industrial Informatics*, vol. 1, no. 2, pp. 97-111, May 2005.

[4] G. Acampora, V. Loia, C. S. Lee, and M. H. Wang, "On the Power of Fuzzy Markup Language," Springer-Verlag, Germany, Jan. 2013.

[5] IEEE Standards Association, "1855-2016 - IEEE Standard for Fuzzy Markup Language," May 2016, [Online] Available: http://ieeexplore.ieee.org/document/7479441/?arnumber=7479441&filter=AND(p_Publication_Number:7479439).

[6] G. Acampora and A. Vitiello, "Interoperable neuro-fuzzy services for emotion-aware ambient intelligence," *Neurocomputing*, vol. 122, no. 3-2, Dec. 2013.

[7] G. Acampora, C. S. Lee, A. Vitiello, and M. H. Wang, "Evaluating cardiac health through semantic soft computing techniques," *Soft Computing*, vol.16, no. 7, pp. 1165-1181, Jul. 2012.

[8] C. S. Lee, M. H. Wang, Y. J. Chen, H. Hagras, M. J. Wu, and O. Teytaud, "Genetic fuzzy markup language for game of NoGo," *Knowledge-Based Systems*, vol. 34, pp. 64-80, Oct. 2012.

[9] C. S. Lee, M. H. Wang, H. Hagas, Z. W. Chen, S. T. Lan, S. E. Kuo, H. C. Kuo, and H. H. Cheng, "A novel genetic fuzzy markup language and its application to healthy diet assessment," *International Journal of Uncertainty, Fuzziness, and Knowledge-Based Systems*, vol. 20, no. 2, pp. 247-278, Oct. 2012.

[10] C. S. Lee, M. H. Wang, M. J. Wu, O. Teytaud, and S. J. Yen, "T2 FS-based adaptive linguistic assessment system for semantic analysis and human performance evaluation on game of Go," *IEEE Transactions on Fuzzy Systems*, vol. 23, no. 2, pp. 400-420, Apr. 2015.

[11] C. S. Lee, M. H. Wang, and S. T. Lan, "Adaptive personalized diet linguistic recommendation mechanism based on type-2 fuzzy sets and genetic fuzzy markup language," *IEEE Transactions on Fuzzy Systems*, vol. 23, no. 5, pp. 1777-1802, Oct. 2015.

[12] C. S. Lee, Z. W. Jian, and L. K. Huang, "A fuzzy ontology and its application to news summarization," *IEEE Transactions on Systems, Man and Cybernetics Part B: Cybernetics*, vol. 35, no. 5, pp. 859-880, Oct. 2005.

[13] G. Acampora, M. Reformat, and A. Vitiello, "Extending FML with evolving capabilities through a scripting language approach," *in Proceeding of 2014 IEEE International Conference on Fuzzy Systems (FUZZ-IEEE 2014)*, Beijing, China, Jul. 6-11, 2014, pp. 857-864.

[14] F. Marini and B. Walczak, "Particle swarm optimization (PSO). A tutorial," *Chemometrics and Intelligent Laboratory Systems*, vol. 149, pp. 153-165, Dec. 2015.

[15] C. Karakuzu, F. Karakaya, and M. A. Cavuslu, "FPGA implementation of neuro-fuzzy system with improved PSO learning," *Neural Networks*, vol. 79, pp. 128-140, Jul. 2016.

[16] F. Zhao, Y. Liu, C. Zhang, and J. Wang, "A self-adaptive harmony PSO search algorithm and its performance analysis," *Expert Systems with Applications*, vol. 42, no. 21, pp. 7436-7455, 2015.

[17] R. Martinez-Soto, O. Castillo, and L. T. Aguilar, "Type-1 and type-2 fuzzy logic controller design using a hybrid PSO-GA optimization method," *Information Sciences*, vol. 285, pp. 35-49, Nov. 2014.

[18] P. Songmuang and M. Ueno, "Bees algorithm for construction of multiple test forms in e-testing," *IEEE Transactions on Learning Technologies*, vol. 4, no. 3, pp. 209-221, Jul.-Sep. 2011.

[19] M. H. Wang, C. S. Wang, C. S. Lee, S. W. Lin, and P. H. Hung, "Type-2 fuzzy set construction and application for adaptive student assessment system," *in Proceeding of 2014 IEEE International Conference on Fuzzy Systems (FUZZ-IEEE 2014)*, Beijing, China, Jul. 6-11, 2014, pp. 888-894.

[20] S. E. Embretson and S. P. Reise, Item Response Theory. Taylor & Francis, 2000.

[21] I. H. Witten and E. Frank, Data Mining. New York: Morgan Kaufman, 1999.

[22] H. Swaminathan and J. A. Gifford, "Bayesian estimation in the three-parameter logistic model," *Psychometrika*, vol. 51, no. 4, pp. 589-601, Dec. 1986.

[23] H. Ozyurt, O. Ozyurt, A. Baki, and B. B. Guven, "Integrating computerized adaptive testing into UZWEBMAT: Implementation of individualized assessment module in an e-learning system," *Expert Systems with Applications*, vol. 39, pp. 9837-9847, Aug. 2012.

[24] H. Lu, Y. P. Hu, J. J. Gao, and Kinshuk, "The effects of computer self-efficacy, training satisfaction and test anxiety on attitude and performance in computerized adaptive testing," *Computers & Education*, vol. 100, pp. 45-55, 2016.

[25] M. Uto and M. Ueno, "Item response theory for peer assessment," *IEEE Transactions on learning technologies*," vol. 9, no. 2, pp. 157-170, Apr.-Jun. 2016.

[26] J. Boy, R. A. Rensink, E. Bertini, and J. D. Fekete, "A principled way of assessing visualization literacy," *IEEE Transactions on Visualization and Computer Graphics*, vol. 20, no. 12, pp. 1963-1972, Dec. 2014.

[27] W. T. Tseng, "Measuring English vocabulary size via computerized adaptive testing," *Computers & Education*, vol. 97, pp. 69-85, Jun. 2016.

[28] M. A. Rau, H. E. Bowman, and J. W. Moore, "An adaptive collaboration script for learning with multiple visual representations in chemistry," *Computers & Education*, vol. 109, pp. 38-55, Jun. 2017.

[29] M. H. Wang, C. S. Wang, C. S. Lee, O. Teytaud, J. L. Liu, S. W. Lin, and P. H. Hung, "Item response theory with fuzzy markup language for parameter estimation and validation," *2015 IEEE International Conference on Fuzzy Systems (FUZZ-IEEE 2015)*, Istanbul, Turkey, Aug. 2-5, 2015.

[30] N. Hansen, D. V. Arnold, and A. Auger, "Evolution strategy," 2013, [Online]. Available: https://www.lri.fr/~hansen/es-overview-2014.pdf.

[31] H. G. Beyer and␣␣P. Schwefel, "Evolution strategies-A comprehensive introduction," *Natural Computing*, vol. 1, no. 1, pp. 3-52, 2002.

[32] M. H. Wang, Z. R. Yan, C. S. Lee, P. H. Hung, Y. L. Kuo, H. M. Wang, and B. H. Lin, "Apply fuzzy ontology to CMMI-based ASAP assessment

system," *2010 IEEE World Congress on Computational Intelligence (IEEE WCCI 2010)*, Barcelona, Spain, Jul. 18-23, 2010.

[33] National Center for Education Statistics, USA. (2015). [Online]. Available: http://nces.ed.gov/.

[34] Z. H. Che, "A particle swarm optimization algorithm for solving unbalanced supply chain planning problems," *Applied Soft Computing*, vol. 12, no. 4, pp. 1279-1287, Apr. 2012.

[35] C. S. Lee, M. H. Wang, T. X. Huang, L. C. Chen, Y. C. Huang, S. C. Yang, C. H. Tseng, P. H. Hung, and N. Kubota, "Ontology-based fuzzy markup language agent for student and robot co-learning," Jan. 2018, [Online] Available: https://arxiv.org/abs/1801.08650.

[36] C. S. Lee, M. H. Wang, S. C. Yang, and C. H. Kao, "From T2 FS-based MoGoTW system to DyNaDF for human and machine co-learning on Go," in R. John, H. Hagras, and O. Castillo (editors), *Type-2 Fuzzy Logic and Systems. Studies in Fuzziness and Soft Computing*, vol. 362, Springer, Cham, 2018, pp. 1-24.